\documentclass{article} % For LaTeX2e
\usepackage{iclr2027_conference,times}
\iclrfinalcopy
\usepackage{amsmath,amsfonts,bm}

\def\eqref#1{equation~\ref{#1}}
\def\1{\bm{1}}

\DeclareMathAlphabet{\mathsfit}{\encodingdefault}{\sfdefault}{m}{sl}
\SetMathAlphabet{\mathsfit}{bold}{\encodingdefault}{\sfdefault}{bx}{n}

\DeclareMathOperator*{\argmax}{arg\,max}

\usepackage{hyperref}
\usepackage{url}

\usepackage{booktabs}
\usepackage[colorinlistoftodos,textsize=tiny]{todonotes}
\usepackage[table]{xcolor}
\usepackage{graphicx}
\usepackage{multirow}
\usepackage{bbm}
\usepackage{wrapfig}
\usepackage{enumitem}
\usepackage{caption}

\usepackage{fontawesome5}

\usepackage{booktabs,tabularx,array,amssymb,placeins}

\newcommand{\method}{\textbf{\texttt{EAU}}}

\title{\textbf{\texttt{eau}}: \textbf{E}nvironment-\textbf{A}gent-\textbf{U}ser Co-evolution\\for Training User-Centric Agents}
\author{
Min Cai\textsuperscript{1,*}
\quad
Dan Zhang\textsuperscript{2,*}
\quad
Gao Xing\textsuperscript{3}
\quad
Libin Ma\textsuperscript{3}
\quad
Zhaohui Xu\textsuperscript{4}
\quad
Xi Ye\textsuperscript{1,5}
\\[1ex]
\textsuperscript{1}University of Alberta
\quad
\textsuperscript{2}National University of Singapore
\quad
\textsuperscript{3}Kuaishou Technology
\\
\textsuperscript{4}Independent Researcher
\quad
\textsuperscript{5}Alberta Machine Intelligence Institute (Amii) 
\\[0.5ex]
\textsuperscript{*}Equal contribution
\quad
\texttt{\{mcai8,xi.ye\}@ualberta.ca}
\\[0.5ex]
\quad \quad \quad \quad \quad \quad \quad \quad \quad \quad \quad \  \faGlobe\texttt{\href{eau-meta-agent.github.io}{eau-meta-agent.github.io}}
}

\begin{document}

\maketitle

\begin{abstract}
\vspace{-10px}
% \xiye{Title: I would probably do EAU: environment-agent-user co-evolution for user centric agentic reinforcement learning}

Training LLM agents to interact with users over multiple turns is challenging because the learning signal depends not only on the agent, but also on the user simulator and environment, which jointly shape the agent's experience. Simulated users can be stochastic or inconsistent, while environments may contain unreliable tools, incomplete state coverage, and incorrect verifiers. Prior work mainly addresses the resulting instability through reward shaping or denser supervision, while treating these components as fixed. We instead introduce \method, a meta-agent framework for \textbf{e}nvironment--\textbf{a}gent--\textbf{u}ser co-evolution. During training, \method~monitors trajectories, tool traces, rewards, and training statistics; diagnoses failures and uninformative interactions; and proposes targeted modifications to the user simulator, environment, reward design, and RL configuration. Validated changes are hot-swapped into the ongoing run, allowing the training setup to evolve together with the policy. These adaptations also persist across runs, so later training can build on previously discovered improvements rather than restarting from a fixed setup. We enable this process through an editable symbolic representation of user and environment dynamics, while the agent remains a neural LLM optimized with RL. 
Across three interactive-agent benchmarks and two backbone models, \method~ consistently outperforms prior RL methods, improving the overall score over the strongest baseline by 14.7--15.0\% and increasing successful task completion on ColBench by up to 36.7\%, showing that co-evolving the training system with the policy can substantially improve multi-turn RL.
% Across two interactive-agent benchmarks, \method~consistently yields strong improvements over prior methods, demonstrating that adapting the interaction process itself can substantially improve multi-turn RL training.

\end{abstract}
\begin{figure}[h]
    \centering
    \includegraphics[width=0.99\linewidth]{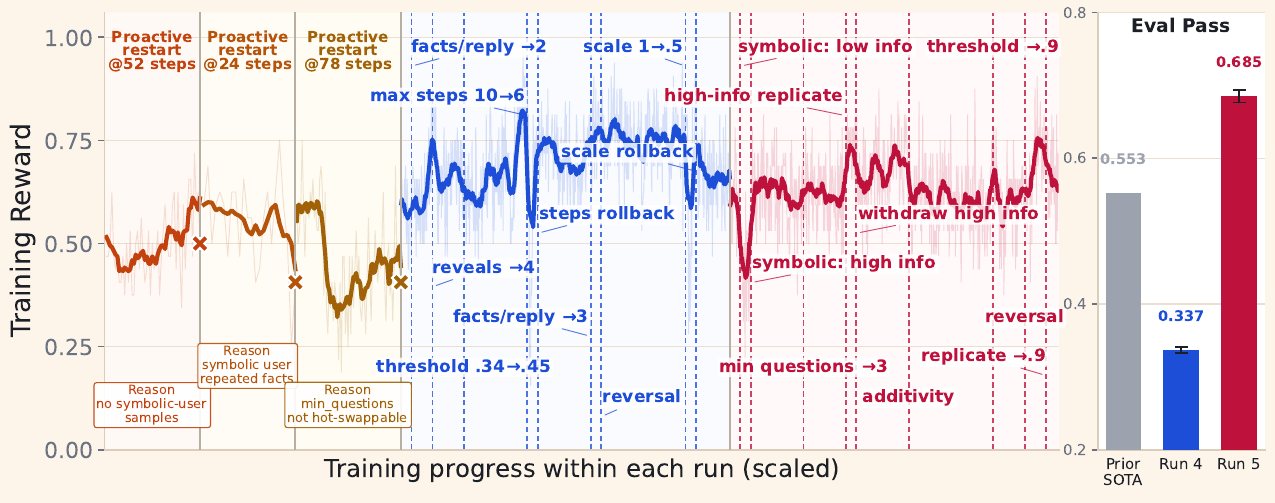}
    \caption{
    Illustration of how our meta-agent, \method, manages RL training runs.  \method~surpasses the prior SOTA on ColBench. \textbf{Left:} For the first three runs, \method~proactively restarts when it identifies issues that can only be resolved with offline patches. \textbf{Middle:} Run \textcolor{blue}{4} and \textcolor{red}{5} displayed in \textcolor{blue}{blue} and \textcolor{red}{red} are two full runs. Dashed lines indicate online, intra-run modifications. Run 4 explored individual modifications through local perturbations, whereas Run 5 used endpoint interventions followed by replication, reversal, and interaction tests.
    % \xiye{Needs reworking}
    % \Dan{Reduce the blank areas at the top and bottom of the entire figure.}
    % With only 5 automated runs \Dan{explain the relationship between the number of runs and the number of steps displayed on the x-axis} on \textit{Qwen3-4B} and subsequently 2 runs on \textit{Qwen2.5-7B}, \method~ achieves the state-of-the-art results on ColBench across 2 models, surpassing the previous SOTA with up to 36.7\% Success Rate (\textbf{Right}). \textbf{Left:} In the first 3 runs of \textit{Qwen3-4B}, \method~identified 3 issues and restarted the experiment \Dan{Indicate the method of representing dashed and solid lines}. 
    % \textbf{Middle:} \method~enables evolution of environment, agent and user during training, exploring different knobs \textit{intra-run} and exploiting experience \textit{inter-run}.
    % \Dan{Right: increase the minimum value of the y-axis, add a hatch to specify different models, and explicitly draw/label the increase ratio.}
    }
    \label{fig:teaser}
    % \vspace{-20px}
\end{figure}
\section{Introduction}

\begin{wrapfigure}[23]{r}{0.35\linewidth}
    \centering
    \includegraphics[width=0.99\linewidth]{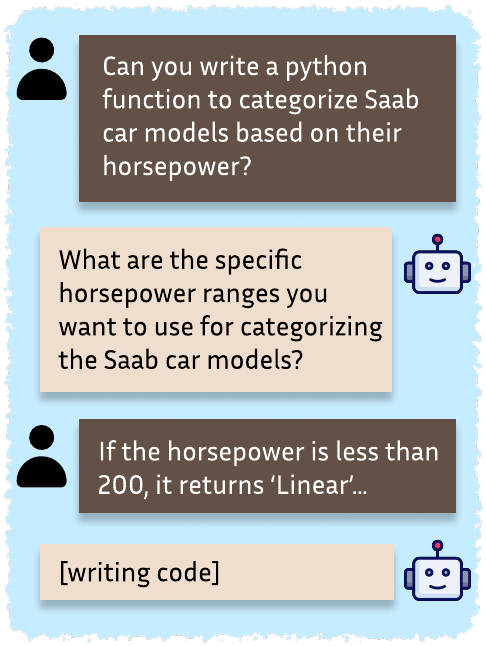}
    \caption{Illustration of a user-centric agentic task. An agent needs to ask for clarifications about a user query.}
    \label{fig:illustration_setting}
\end{wrapfigure}

Large language models are increasingly deployed as agents that collaborate with humans rather than simply execute fully specified tasks. As shown in Figure~\ref{fig:illustration_setting}, in many realistic applications, users can provide incomplete or evolving specifications, requiring an agent to interact over multiple turns: asking clarifying questions, resolving ambiguities, negotiating constraints, and adapting its behavior as new information emerges. This has motivated a growing body of interactive benchmarks in which agents communicate with simulated users while acting in an external 
environment~\citep{yao2024tau,barres2025tau2,qian2025userbench}. Such simulation makes large-scale evaluation and reinforcement learning (RL) possible, but it also introduces a fundamental challenge: \textit{the learning signal is only as reliable as the interaction that produces it}. LLM-based user simulators are stochastic and can behave inconsistently, for example by drifting from their assigned persona, revealing information prematurely, or changing their intent across otherwise similar episodes. At the same time, environments may contain incomplete state coverage, unreliable tool behavior, unreachable tasks, or incorrect verifiers. Moreover, because both components are typically fixed throughout training, the interactions they generate need not remain informative as the policy evolves. Together, these issues can make rewards highly variable or systematically misleading, substantially destabilizing RL for interactive agents.

Prior work has largely addressed this instability by improving the supervision presented to the policy, for example through reward shaping that credits partial progress~\citep{qian2025userrl} or auxiliary supervision that provides denser learning signals~\citep{kong2026infopo,wang2025igpo}. These approaches can make better use of a noisy signal, but they leave the sources of that signal---the user simulator and the environment---largely unchanged. We argue that this misses an important root cause of instability. If a tool silently fails for a particular argument, a verifier incorrectly rejects a successful trajectory, or a simulated user reveals information that the agent was supposed to elicit, modifying the downstream reward cannot recover the intended interaction. More generally, as the policy improves, a fixed simulator and environment may continue producing trajectories that are redundant, uninformative, or poorly targeted to the policy's current weaknesses. We therefore propose shifting part of the intervention upstream: rather than treating the user simulator and environment as fixed infrastructure and optimizing only the policy's learning signal, we continually diagnose and revise the interaction process itself during training.
% \Dan{add the ref of Figure 1 and Figure 2.}

To realize this idea, we introduce \method, a meta-agent that continuously supervises RL training and improves the components that generate its learning experience. Throughout training, \method~observes rollout transcripts, tool traces, rewards, and aggregate training statistics, and diagnoses whether problematic or uninformative trajectories arise from the policy, the user simulator, the environment, or the training configuration. As shown in Figure~\ref{fig:teaser}, it proposes targeted interventions based on this diagnosis, including correcting environment or verifier failures, refining the simulator's persona and information-disclosure behavior, and adjusting reward terms or training hyperparameters. Importantly, these interventions are not limited to fixing errors. As the policy improves, \method~can also adapt the simulator and environment to expose unresolved weaknesses and generate more informative training trajectories. Candidate modifications are validated by a bounded integrity gate before being accepted and hot-swapped into the ongoing training loop, allowing the training system itself to evolve without repeatedly restarting from scratch.

This produces a different view of interactive-agent training: rather than optimizing a policy against a fixed user and environment, \method~enables the training system to adapt alongside policy learning. Instead of prescribing a fixed sequence of component updates, the meta-agent diagnoses emerging problems and determines what to modify and when to intervene. Simulator revisions, reward and curriculum adjustments, hyperparameter changes, and implementation repairs are therefore part of the intervention space. The contribution lies in enabling and organizing this adaptive process during and across training runs.

We evaluate \method~on three interactive-agent benchmarks and two backbone models. Across these settings, \method~discovers interpretable modifications to both simulators and environments and improves final agent performance by up to 36.7\% over the previous state of the art, while our analysis shows how the three components evolve together throughout training.

To summarize, our contributions are:
\begin{itemize}[noitemsep,topsep=0pt,leftmargin=*]
\item We identify fixed user simulators and environments as a key bottleneck in interactive RL.
\item We introduce \method, a meta-agent that diagnoses and updates the training system online.
\item We demonstrate improved agent performance on three user-centric benchmarks, with targeted analyses of symbolic-user mixing, intra-run intervention, and alternative training algorithms and meta-agent implementations.
\end{itemize}

\section{Related Work}
\label{sec:related_work}
\paragraph{User Centric Agentic Reinforcement Learning.}
User-centric tasks~\citep{yao2024tau,barres2025tau2,qian2025userbench} focus on interacting and fulfilling the requirement of a simulated user under multi-turn conversations. Different from previous multi-turn Reinforcement Learning~\citep{jimenez2024swebench,xi2025agentgymrl,zhang2026cm2reinforcementlearningchecklist} that solely interacts with a fixed deterministic environment, user-centric RL~\citep{qian2025userrl,kong2026infopo,yu2026frozensimulatorenoughsimulator} poses an additional challenge: the introduction of LLM user simulator may cause several issues such as stochastic state transition, inconsistent behaviors and collapse in training due to overfitting a fixed behavior~\citep{yu2026frozensimulatorenoughsimulator}. Recent methods mainly focus on improving supervision signal presented to the policy~\citep{zhou2025sweet,kong2026infopo}. \cite{yu2026frozensimulatorenoughsimulator} co-train the user to avoid overfitting. Our work identifies that issues may lie in all three parts, i.e., RL training dynamics, user simulation and environment, and we endeavor to address all at once with a meta agent, \method.

\paragraph{Meta-Agents and Automated Optimization of Training Components.}
A growing body of work uses one LLM to optimize another system or the system in which it operates~\citep{hu2025adas,lee2026metaharness,zhang2026darwingodelmachineopenended,liu2026adaptiveautoharnesssustainedselfimprovement,chen2026learningselfevolve,surina2025algorithmdiscoveryllmsevolutionary}. \method~differs mainly in that the meta-agent acts \emph{within} a run: modifications are validated and hot-swapped into a live training loop rather than applied between runs or before training begins.

\paragraph{Co-evolution of Agent and its World.}
More recently, there's a trend in improving the world, e.g., environment and user simulator, that an agent is interacting with. While most of them prepare the worlds before training~\citep{huang2026envharness,tu2026scaleenvscalingenvironmentsynthesis}, some begin to discover how the world can evolve along with the agent training~\citep{yu2026frozensimulatorenoughsimulator,kim2026frognanotraining4bcoding,yang2026coevolvetrainingllmagents}. Unlike prior methods that evolve the world through predefined update rules or workflows, \method~ can flexibly determine what to modify, when to intervene, and how to adapt the training system online.

\section{Problem Setup}
\begin{figure}[t]
    \centering
    \includegraphics[width=0.99\linewidth]{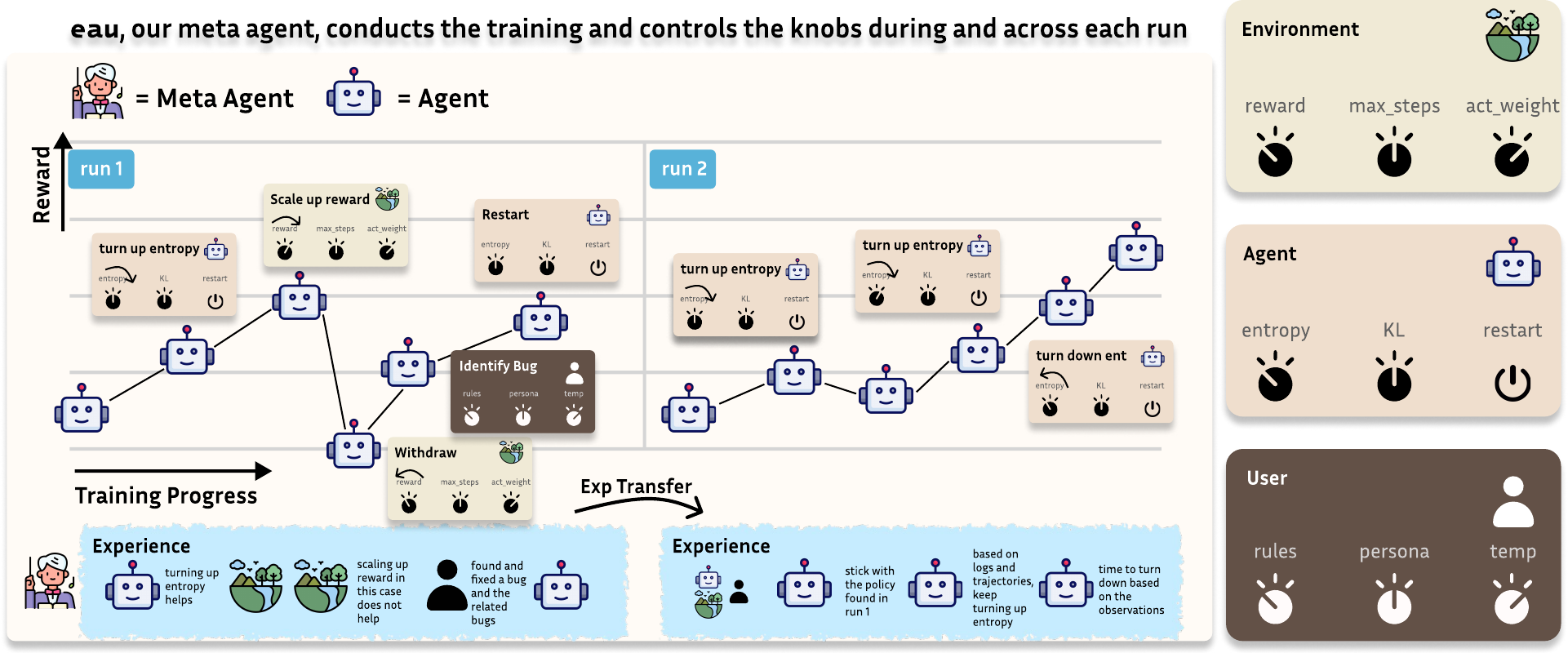}
    \caption{Illustration of how \method, our meta agent, explores, learns, and absorbs experience online during and across each run. Reading logs and trajectories, \method~explores different strategies online by controlling the \textit{knobs} of each module, learns which strategies help and which do not. Experience is brought across different runs, serving as a solid foundation for continual exploration and learning.}
    \label{fig:eau_illustration}
\end{figure}

We focus on user-centric agentic tasks, where an agent interacts with both an environment through tool calls and a (simulated) user over multiple turns. The goal is to accomplish a task while satisfying the user's intent, which may be underspecified at the outset and gradually refined through the interaction.

We formulate each task in our setting as a partially observable Markov decision process (POMDP). Formally, the environment for a task is $\mathcal{E} = \langle \mathcal{S},\mathcal{A},\mathcal{T},\mathcal{R},\mathcal{Z},\mathcal{O}\rangle$ with state space $\mathcal{S}$, action space $\mathcal{A}$, transition function $\mathcal{T}: \mathcal{S} \times \mathcal{A} \rightarrow \Delta(\mathcal{S} \times \mathcal{O})$, reward function $\mathcal{R}: \mathcal{S} \times \mathcal{A} \rightarrow [0, 1]$, observation space $\mathcal{Z}$, and a function $\mathcal{O}: \mathcal{S} \rightarrow \Delta(\mathcal{Z})$ that maps states to observations. $\Delta$ here refers to the stochasticity in transitions and observations. We consider each task to have a simulated user as a part of the environment, therefore an agent needs to interact with the environment by tool calling and the user simulator by conversations, i.e., $\mathcal{S}=\mathcal{S}_{\textrm{tool}} \otimes \mathcal{S}_{\textrm{user}} \otimes \mathcal{S}_\textrm{query}$, where $\mathcal{S}_{\textrm{query}}$ can be considered as the first state. And we have $\mathcal{A}=\mathcal{A}_{\textrm{tool}}\cup\mathcal{A}_{\textrm{user}}$ , $\mathcal{Z}=\mathcal{Z}_{\textrm{tool}} \cup \mathcal{Z}_{\textrm{user}}$. Specifically, for transition $\mathcal{T}$, interactions with the environment through tool calls are deterministic, while interactions with simulated users are stochastic. Thus, we have $\mathcal{T}_{\textrm{tool}}:(s_{\textrm{tool}}, a_{\textrm{tool}}) \mapsto (s'_{\textrm{tool}}, z_{\textrm{tool}})$ and $\mathcal{T}_{\textrm{user}}:(s_{\textrm{user}}, a_{\textrm{user}}) \rightarrow (\mathcal{S}_{\textrm{user}}, \mathcal{Z}_{\textrm{user}})$.

Let $\pi_{\theta}$ be the policy (agent), our goal is to maximize the number of successful tasks, i.e., $\max_{\pi} \mathbb{E}_{\tau \sim \pi} \left[ \sum_{t=0}^{T} R(s_t, a_t) \,\middle|\, \textrm{query} \right]$. 
To achieve this, a policy needs to first interact, negotiate, and even persuade the user, while at the same time understanding users' intents and calling the correct tools. 
Thus, we term it a bi-level optimization:
\begin{align}
\max_{\theta} \; \mathbb{E}_{\tau \sim \pi_{\theta}} \left[ R_{\text{task}}(\tau, u) \right] 
&& \text{s.t.} && 
\pi_{\theta} = \argmax_{\pi} \; \mathbb{E}_{\tau' \sim \pi} \left[ \sum_{t=0}^{T-1} \gamma^t r_{\text{conv}}(s_t, a_t) \right],\label{eq:goal}
\end{align}
where $R_{\textrm{task}}$ refers to rewards related to task success, and $r_{\textrm{conv}}$ refers to rewards related to conversations specifically designed for each environment. Notably, $R_{\textrm{task}}$ is usually verifiable yet sparse while $r_{\textrm{conv}}$ can be dense but noisy.

\section{Method: Environment-Agent-User Co-Evolution}
\label{sec:method}
% Core problem:
% 1) user and enviroment is stohastic, user-simulator
% 2) not reliable: not follow task instruction (reward is not reliable, not following format, or answer leakage).

In this section, we introduce \method~(Environment-Agent-User Co-Evolution) for user-centric agentic RL. Recall that agentic RL with simulated users presents several challenges.
First, sparse rewards and stochastic interactions can make learning difficult.
Second, user simulators may provide unreliable rewards, deviate from instructions, or leak answers during conversations.

\subsection{Method Overview}
The core idea of \method~is to couple policy learning with adaptation of the system that generates its training experience. The agent is updated through RL, while a meta-agent uses training trajectories and optimization statistics to revise the user simulator, environment, and configurable training dynamics. These revisions change the interactions and learning signals encountered by subsequent policy updates, closing the feedback loop between policy behavior and training-system adaptation.

We use \emph{co-evolution} to denote this meta-agent-mediated, asymmetric co-adaptation. It does not require the environment and user simulator to be independently learned models: their adaptation is implemented through programmatic interventions. As illustrated in Figure~\ref{fig:eau_illustration}, the meta-agent observes training feedback, selects interventions through editable interfaces, and accumulates experience for subsequent decisions. The following subsections describe the user-, environment-, and agent-side interventions, followed by their coordination within and across training runs.
% \cm{haven't adjusted the followings into this ``knob'' narrative}\xytodo{This is much better. I am mainly just polishing things}

\subsection{Evolving User}
% how we evolve user: keep updating rules, epsilon-symbolic, evolution: means update regex rules, determistic functions that takes parsed agent outputs and return a value (tiny illustrative thing)
% rule and prompt 

\begin{wrapfigure}[17]{r}{0.55\linewidth}
    \centering
    \includegraphics[width=0.99\linewidth]{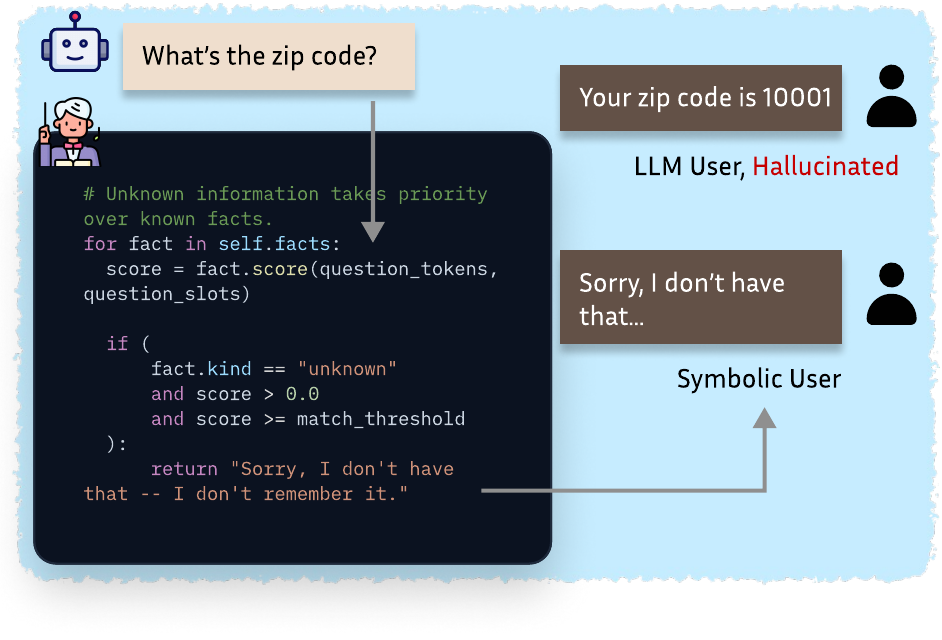}
    \caption{Illustration of symbolic user simulator.}
    \label{fig:illustration_user_evo}
\end{wrapfigure}
LLM-simulated users are inherently stochastic and may imperfectly follow its specification, for example by revealing information that should remain hidden or by improvising answers to questions unsupported by the task. To provide a more controllable signal, we introduce a \emph{symbolic user simulator}, $u_{\mathrm{sym}}$, which consists of a set of explicit rules that deterministically map interaction states, obtained from parsed agent outputs, to user responses. Figure~\ref{fig:illustration_user_evo} illustrates an example of such a rule. These rules provide precise control over what information is revealed and under which conditions. However, designing and updating them manually throughout training is intricate and labor-intensive. We therefore expose the symbolic rules as tunable \textit{knobs}, allowing the meta-agent to revise them in response to the agent's evolving behavior.

The symbolic and LLM simulators offer complementary strengths: the former provides controllability and determinism, while the latter offers richer and more natural interactions. We combine them through an $\epsilon$-symbolic strategy:

\begin{align} u_\epsilon := \begin{cases} u_{\mathrm{sym}}, & \text{with probability } \epsilon,\\ u_{\mathrm{LLM}}, & \text{with probability } 1-\epsilon, \end{cases} \qquad \epsilon \in [0,1]. \end{align}

The mixing coefficient $\epsilon$ thus controls the balance between structured, deterministic supervision and flexible LLM-based interaction, and can itself be adapted by the meta-agent during training.

Beyond the symbolic rules and mixture coefficient $\epsilon$, \method~can also evolve the LLM user by tuning controllable generation parameters, including temperature and persona-specific prompt instructions. Together, these mechanisms allow the user side of the training loop to adapt alongside the policy rather than remaining fixed throughout training.

% Enviroment: 
% * mainly changed rewards or add enviromental rewards, intermediate
\begin{wrapfigure}[13]{r}{0.5\linewidth}
    \centering
    \includegraphics[width=0.99\linewidth]{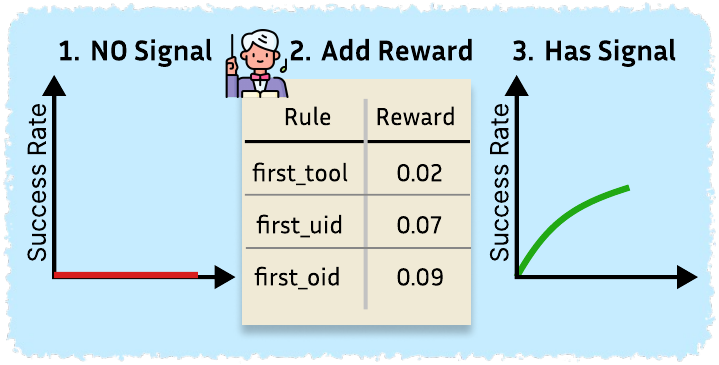}
    \caption{Illustration of environment evolution by introducing additional rewards.}
    \label{fig:illustration_env_evo}
\end{wrapfigure}
\subsection{Evolving Environment}
Environment-side adaptation in \method~modifies the training feedback and supported interaction conditions, rather than learning a new world model. In our implementation, this primarily involves adding or revising intermediate reward signals in $R$, together with benchmark-specific interaction constraints such as episode length. Recall that user-centric agentic tasks typically involve multi-turn interactions, where verifiable rewards are often sparse, such as task success. As a result, the agent may struggle to explore effectively when it rarely reaches successful outcomes. When such behavior is observed, \method~can introduce intermediate reward rules that identify and reward states indicative of progress toward task completion.
An illustrative example is shown in Figure~\ref{fig:illustration_env_evo}.
% \cm{do we need an example here?} 
% \xiye{Good to have a little bit sense of concrentess. lets chat a bit for this}

\method~can also revise existing manually designed rewards when they induce undesirable behaviors or divert the agent from the final objective. By dynamically adding and modifying intermediate rewards, \method~provides more informative learning signals while also accumulating experience about which reward designs facilitate or hinder learning. This experience can in turn guide future environment evolution.

% pararaph: On-the-fly hyper-paramters managing, and its advantages, we allow on the fly hyp-tuning
\subsection{Evolving Agent}
The agent naturally evolves through gradient updates during RL training. We keep the underlying \emph{training algorithm fixed} and let the meta-agent continuously monitor the optimization process and intervene in its configurable training dynamics when necessary. 

\paragraph{Training Algorithm.}
\label{sec}
We use and fix Proximal Policy Optimization~\citep[PPO]{schulman2017ppo} as the RL algorithm. Recent work on LLM RL mainly use critic-free methods, such as GRPO~\citep{shao2024grpo}, for low training cost and unbiased advantage estimation. However, current critic-free methods suffer from the sparse credit assignment, where intermediate rewards are accumulated to a single trajectory-level reward~\citep{qian2025userrl}, or are not applicable to user-centric tasks where the same states are hard to replicate for group advantage estimation~\citep{feng2026group}.

% \xiye{Is this standard? do we have any adaptions?}
We therefore use a learned critic $V_\phi$, parameterized by $\phi$, with Generalized Advantage Estimation (GAE). The critic is optimized by minimizing the value error $\mathcal{L}^{\mathrm{VF}}_\phi = \mathbb{E}[(V_\phi(q, y_{<t}) - R)^2]$, where $R$ is the cumulative reward. Both critic and policy are restricted to action tokens: the critic is evaluated only where the policy acts, and GAE is computed across actions rather than across all tokens, so the advantage of a turn is not diluted by the observation text surrounding it~\citep{hou2026single}. Formally, the advantage $\hat{A}^{\mathrm{GAE}}_t$ is computed as:
\begin{align}
    \hat{A}^{\mathrm{GAE}}_t = \sum_{l=0}^{|y|-t-1}(\gamma\lambda)^l \delta_{t+l},
\end{align}
where $\delta_t = r_t + \gamma V_\phi(s_{t+1}) - V_\phi(s_t)$, and $s_t$ refers to the $t$-th action token.

\paragraph{Monitoring Training}
During training, the meta agent is allowed to tune the hyper-parameters such as entropy coefficient and KL coefficient, based on training logs, including metrics like estimated rollout entropy and log likelihood. It can also read the rollout trajectories and the interactions with the environment and the user.
% \xiye{During the training run, the meta agent is allowed to modify hyper-peratermers such as entropy, ... KL coeffeciient, based on training logs (loss and validation performance entropy, response length per steps).}

% \xiye{we can discuss intra and inter run together}

\subsection{Intra- and Inter-run Intervention}

A notable design choice of \method~is to support flexible intervention at two timescales: both \textit{intra-run} and \textit{inter-run}. As illustrated in Figure~\ref{fig:eau_illustration}, for intra-run intervention, updates can be hot-swapped into the training loop without restarting from scratch. For example, \method~can revise symbolic-user rules, add or modify environmental rewards, and adjust RL hyperparameters during training.

It can also decide when a restart (inter-run) is necessary, such as after an irreversible failure, or resume from an intermediate checkpoint. Inter-run intervention, in contrast, enables \method~to consolidate observations from completed runs and use them to revise the setup for subsequent training.

% \xiye{Is it the case that mot prior design are inter run only? If so we should highlight that and cite some work}
% \xiye{need a sentence unlike prominent paradigmn only allows inter-run (full restart).cite xxxxx.....}
Unlike previous paradigm that only allows inter-run modifications~\citep{chen2026evotrainer,xia2026ai,liu2024large,yamada2025ai,ferreira2026can}, by granting this flexibility, \method~can explore different operations on-the-fly, which is more efficient than inter-run interventions alone. Therefore, it can find out more effective operations based on the intra- and inter-run experience.

\subsection{Instantiation of Meta-Agent} 
We instantiate our meta agent with coding agents, such as Claude Code and Codex, following recent works on meta agent~\citep{lee2026metaharness}. The meta agent has access to a local filesystem which includes the source code, training logs, etc. It reads the logs with a fixed interval during training, and decides which part to modify for environment-user-agent evolution. A list of \textit{knobs} to tune is prepared at the beginning, but \method~can also modify the list if necessary. It is also asked to maintain a experience pool as a directory of documents in the filesystem. Before each modification, it may reference to the experience pool and whenever there's a new finding, it updates the pool accordingly.

Unlike previous methods that allow the meta-agent to collect feedback on an evaluation/test set~\citep{panfilov2026claudini,lee2026metaharness,xia2026ai}, \method~ uses only training-side feedback to guide its interventions. The workflow described above is implemented as a skill comprising instructions and supporting scripts.

\section{Experiments}
\label{sec:experiment}
\begin{table*}[t]
\centering
\caption{Comparison across UserGym and ColBench on \textit{Qwen3-4B} and
\textit{Qwen2.5-7B-Instruct}. The best and second-best scores in each category are highlighted in \textbf{bold} and \underline{underlined} respectively. Standard deviations across 3 repeated evaluations are shown in the subscripts.}
\label{tab:main_results}
\scriptsize
\setlength{\tabcolsep}{3pt}
\renewcommand{\arraystretch}{1.08}

% -------------------------------------------------
% Color palette
% -------------------------------------------------
\definecolor{sectiongray}{RGB}{244,245,247}
\definecolor{oursyellow}{RGB}{252,247,229}

% Vertical separators:
% UserGym | ColBench | Overall
\begin{tabular}{
    ll
    cccccccc
    !{\vrule width 0.4pt}
    ccc
    !{\vrule width 0.4pt}
    c
}
\toprule

\multirow{3}{*}{\textbf{Type}} &
\multirow{3}{*}{\textbf{Method}} &
\multicolumn{8}{c!{\vrule width 0.4pt}}{\textbf{UserGym}} &
\multicolumn{3}{c!{\vrule width 0.4pt}}{\textbf{ColBench}} &
\multirow{3}{*}{\textbf{Overall}} \\

& &
\multicolumn{5}{c}{\textit{ID}} &
\multicolumn{2}{c}{\textit{OOD}} &
\multirow{2}{*}{Avg.} &
\multirow{2}{*}{Pass} &
\multirow{2}{*}{Succ.} &
\multirow{2}{*}{Avg.} &
\\[-0.3ex]

% Two separate subgroup lines, without breaking vertical separators
\noalign{\vskip-\aboverulesep}
\cmidrule(lr){3-7}
\cmidrule(lr){8-9}
\noalign{\vskip-\belowrulesep}

& &
Travel & Func. & Pers. & Tau & Turtle &
Inten. & Tele. &
& & & & \\

\midrule

% =================================================
% Closed-source
% =================================================
\rowcolor{sectiongray}
\multicolumn{14}{c}{\textit{Closed-Source Models}} \\

Prompting & Gemini-3-Flash &
0.574 & 0.423 & 0.695 & 0.167 & 0.153 &
1.718 & 0.829 & 0.651 &
0.515 & 0.382 & 0.449 & 0.606 \\

Prompting & GPT-4.1 &
0.554 & 0.051 & 0.599 & 0.109 & 0.267 &
1.867 & 0.732 & 0.597 &
0.529 & 0.403 & 0.466 & 0.568 \\

Prompting & GPT-4o-mini &
0.596 & 0.089 & 0.683 & 0.091 & 0.117 &
1.277 & 0.512 & 0.481 &
0.463 & 0.342 & 0.403 & 0.463 \\

\midrule

% =================================================
% Qwen3-4B
% =================================================
\rowcolor{sectiongray}
\multicolumn{14}{c}{\textit{Qwen3-4B}} \\

Prompting & Qwen3 &
0.277 & 0.026 & 0.452 & 0.006 & 0.071 &
1.660 & 0.488 & 0.426 &
0.272 & 0.153 & 0.213 & 0.378 \\

Prompting & ReAct &
0.279 & 0.053 & 0.476 & 0.005 & 0.147 &
1.782 & 0.454 & 0.457 &
0.269 & 0.145 & 0.207 & 0.401 \\

Prompting & Reflexion &
0.291 & 0.045 & 0.474 & 0.035 & 0.124 &
1.717 & 0.501 & 0.455 &
0.303 & 0.184 & 0.244 & 0.408 \\

RL Training & RAGEN &
0.477 & 0.104 & 0.511 & 0.006 & 0.117 &
1.572 & 0.488 & 0.468 &
0.479 & 0.361 & 0.420 & 0.457 \\

RL Training & Search-R1 &
0.482 & 0.103 & 0.425 & 0.059 & 0.113 &
1.711 & 0.412 & 0.472 &
0.467 & 0.355 & 0.411 & 0.459 \\

RL Training & UserRL &
0.538 & 0.095 & 0.507 & 0.053 & 0.121 &
1.732 & \underline{0.520} & 0.509 &
0.468 & 0.342 & 0.405 & 0.486 \\

RL Training & InfoPO &
\underline{0.589} & \underline{0.115} & 0.556 & \underline{0.097} & 0.154 &
1.862 & \textbf{0.542} & \underline{0.559} &
\underline{0.553} & \underline{0.439} & \underline{0.496} & \underline{0.545} \\

\cmidrule(lr){1-14}

RL Training & PPO &
0.468 & 0.051 & \underline{0.560} & 0.006 & \underline{0.180} &
\underline{1.880} & 0.512 & 0.522 &
0.284\textsubscript{$\pm$0.046} &
0.200\textsubscript{$\pm$0.044} &
0.242 & 0.462 \\

% RL Training & Meta-Agent &
% 0.573 & 0.115 & 0.540 & 0.103 & 0.232 &
% 1.965 & 0.488 & 0.574 &
% 0.317\textsubscript{$\pm$0.038} &
% 0.207\textsubscript{$\pm$0.035} &
% 0.262 & 0.505 \\

\rowcolor{oursyellow}
RL Training & \textbf{Ours} &
\textbf{0.615} & \textbf{0.186} & \textbf{0.726} &
\textbf{0.109} & \textbf{0.188} &
\textbf{2.023} & 0.512 &
\textbf{0.623} &
\textbf{0.685\textsubscript{$\pm$0.009}} &
\textbf{0.600\textsubscript{$\pm$0.015}} &
\textbf{0.643} &
\textbf{0.627} \\

\midrule

% =================================================
% Qwen2.5-7B-Instruct
% =================================================
\rowcolor{sectiongray}
\multicolumn{14}{c}{\textit{Qwen2.5-7B-Instruct}} \\

Prompting & Qwen2.5 &
0.441 & 0.026 & 0.289 & 0.000 & 0.062 &
1.254 & 0.292 & 0.338 &
0.242 & 0.140 & 0.191 & 0.305 \\

Prompting & ReAct &
0.452 & 0.064 & 0.325 & 0.037 & 0.078 &
1.378 & 0.316 & 0.379 &
0.238 & 0.135 & 0.187 & 0.336 \\

Prompting & Reflexion &
0.445 & 0.052 & 0.312 & 0.022 & 0.074 &
1.320 & 0.305 & 0.361 &
0.276 & 0.168 & 0.222 & 0.330 \\

RL Training & RAGEN &
0.538 & 0.124 & 0.548 & 0.000 & 0.148 &
1.815 & 0.416 & 0.513 &
0.449 & 0.348 & 0.399 & 0.487 \\

RL Training & Search-R1 &
0.565 & 0.113 & 0.412 & 0.043 & 0.154 &
1.805 & 0.416 & 0.501 &
0.457 & 0.352 & 0.405 & 0.480 \\

RL Training & UserRL &
0.546 & 0.115 & 0.444 & \underline{0.048} & 0.152 &
1.826 & 0.424 & 0.508 &
0.436 & 0.327 & 0.382 & 0.480 \\

RL Training & InfoPO &
0.588 & 0.167 & 0.535 & \textbf{0.091} & \textbf{0.178} &
\underline{1.892} & \underline{0.488} & 0.563 &
\underline{0.534} & \underline{0.426} & \underline{0.480} & \underline{0.544} \\

\cmidrule(lr){1-14}

RL Training & PPO &
\textbf{0.596} & \underline{0.229} & \underline{0.651} & 0.018 & 0.135 &
1.860 & \underline{0.488} & \underline{0.568} &
0.356\textsubscript{$\pm$0.012} &
0.240\textsubscript{$\pm$0.000} &
0.298 & 0.508 \\

% RL Training & Meta-Agent &
% 0.532 & 0.261 & 0.571 & 0.024 & 0.149 &
% 1.558 & 0.488 & 0.512 &
% 0.415\textsubscript{$\pm$0.007} &
% 0.277\textsubscript{$\pm$0.006} &
% 0.346 & 0.475 \\

\rowcolor{oursyellow}
RL Training & \textbf{Ours} &
\underline{0.595} & \textbf{0.333} & \textbf{0.730} &
\underline{0.048} & \underline{0.170} &
\textbf{2.035} & \textbf{0.512} &
\textbf{0.632} &
\textbf{0.659\textsubscript{$\pm$0.017}} &
\textbf{0.533\textsubscript{$\pm$0.018}} &
\textbf{0.596} &
\textbf{0.615} \\

\bottomrule
\end{tabular}
\end{table*}

% RL Training & \textbf{Ours (iter 1, CoSER)}& 0.448& 0.191& 1.028& 0.091& 0.900& 1.823& 0.781& 0.752& - & - & - & - \\
% RL Training & \textbf{Ours (iter 1, 4omini)}& 0.494& 0.169& 0.694& 0.061& 0.209& 1.710& 0.585& 0.560& - & - & - & - \\

\subsection{Evaluation Protocol}
We evaluate \method~on three user-centric benchmarks. Experiments on UserGym and ColBench use two policy backbones, \textit{Qwen3-4B}~\citep{yang2025qwen3} and \textit{Qwen2.5-7B-Instruct}~\citep{qwen2025qwen25technicalreport}, while the $\tau^2$-bench experiments use \textit{Qwen3-4B}. We follow the benchmark evaluation protocol of \citet{kong2026infopo}.

\paragraph{Benchmarks.}
We consider \textbf{ColBench}~\citep{zhou2025sweet}, \textbf{UserGym}~\citep{qian2025userrl}, and \textbf{$\tau^2$-bench}~\citep{barres2025tau2}. These benchmarks require agents to uncover hidden information through conversation and tool calls to fulfill users' requests. Benchmark details are provided in Appendix~\ref{sec:appendix_experiment_details}.

\paragraph{Baselines.}
We compare against prompting methods (ReAct~\citep{yao2022react} and Reflexion~\citep{shinn2023reflexion}) and RL training methods (RAGEN~\citep{wang2025ragen}, Search-R1~\citep{jin2025search}, UserRL~\citep{qian2025userrl}, and InfoPO~\citep{kong2026infopo}). We also include vanilla PPO. For UserGym and ColBench, baseline results other than PPO are taken from \citet{kong2026infopo}. A generic meta-agent baseline is additionally examined in our ablations.

\paragraph{User simulator.}
The default training-time LLM simulator is \textit{CoSER}~\citep{wang2025coser}, a fine-tuned \textit{Llama-3.1-8B} model. This choice is motivated by its competitive user-simulation performance reported by \citet{zhou2026mindsim2realgapuser}. During training, \method~may replace the simulator when justified by the training logs. For evaluation, we use \texttt{GPT-4o-mini}, following \citet{kong2026infopo}.

\paragraph{Evaluation and model selection.}
Evaluation is performed after each completed training run. We report the final run and its final checkpoint for each experiment, without using final-test feedback for selection. Evaluation instances, trajectories, scores, and derived summaries are not used to propose or accept interventions, update the experience pool, or guide subsequent training runs. Training-side modifications do not alter the evaluation task set, user-simulator configuration, interaction limits, or scoring rules.

\subsection{Implementation Details}

\paragraph{Meta-agent backbone.}
Unless otherwise specified, we instantiate \method~with \texttt{Claude Code} using \textit{Claude Opus-5}. Table~\ref{tab:ablation_training_alg_backbone} examines an alternative meta-agent implementation.

\paragraph{Training and search budgets.}
The default episode limits are $16$ turns on UserGym, $10$ on ColBench, and $50$ on $\tau^2$-bench, with a per-turn generation budget of $1024$ tokens and per-episode rollout budgets of $8192$, $8192$, and $16384$ tokens, respectively. For each experiment, we cap the number of inter-run iterations at $5$, including proactive restarts. Remaining optimization settings are provided in Appendix~\ref{sec:additional_implementation_details}. These limits specify the rollout and search setup, rather than a matched end-to-end compute budget; the latter would also need to account for meta-agent inference and all attempted training runs.

\subsection{Main Results}
\begin{wraptable}[10]{r}{0.43\textwidth}
  \centering
  \caption{Results on $\tau^2$-bench (\textit{Qwen3-4B}).}
  \label{tab:tau2bench_results}
  \small
  \setlength{\tabcolsep}{4pt}
  \begin{tabular}{@{}lcccc@{}}
    \toprule
    \multicolumn{1}{c}{Method} & Retail & Airline & Telecom & Avg. \\
    \midrule
  UserRL & 16.9 & 7.5 & 13.8 & 13.8 \\
  InfoPO & \textbf{24.4} & 10.0 & 15.6 & 18.0 \\
  \midrule
  PPO & 16.3 & 23.8 & 18.1 & 18.5 \\
  \method & 18.1 & \textbf{28.8} & \textbf{21.3} & \textbf{21.5} \\
    \bottomrule
  \end{tabular}
\end{wraptable}

As shown in Table~\ref{tab:main_results}, \method~achieves the highest overall score among the evaluated RL methods on both policy backbones. Compared with InfoPO, the strongest prior RL baseline by overall score, \method~improves from $0.545$ to $\mathbf{0.627}$ on \textit{Qwen3-4B} and from $0.544$ to $\mathbf{0.615}$ on \textit{Qwen2.5-7B-Instruct}, corresponding to relative gains of $15.0\%$ and $14.7\%$, respectively. With a $4$B policy backbone, \method~also achieves a higher overall score than the evaluated closed-source models, including Gemini-3-Flash ($0.606$) and GPT-4.1 ($0.568$). These results show consistent overall improvements on UserGym and ColBench across the two tested Qwen backbones, which differ in model size and generation.

\paragraph{UserGym.}
RL training generally provides larger gains than the tested prompting methods. Relative to InfoPO, \method~raises the average UserGym score from $0.559$ to $\mathbf{0.622}$ on \textit{Qwen3-4B} ($+11.3\%$) and from $0.563$ to $\mathbf{0.620}$ on \textit{Qwen2.5-7B-Instruct} ($+12.3\%$). On \textit{Qwen3-4B}, \method~outperforms the best RL baseline in six of the seven evaluated environments, with particularly large gains on Func., Pers., and Inten. Inten.\ and Tele.\ are held out during training and therefore assess cross-environment generalization. On Inten., \method~reaches $\mathbf{2.023}$, exceeding InfoPO ($1.862$) and the evaluated closed-source models. On Tele., it scores $0.512$, slightly below InfoPO ($0.542$). These results provide evidence of transfer to unseen environments, although the gains are not uniform across them.

\begin{wraptable}[9]{r}{0.43\textwidth}
  \centering
  \small
  \caption{Ablation on the user simulator. Results are on ColBench with \textit{Qwen3-4B}.}
  \label{tab:ablation_user}
  \begin{tabular}{lcc}
    \toprule
    Variant & Pass mean & Succ. mean \\
    \midrule
    LLM      & \(0.284\textsubscript{$\pm$ 0.046}\) & \(0.200 \textsubscript{$\pm$0.044}\) \\
    Symbolic & \(0.229\textsubscript{$\pm$0.004}\) & \(0.177\textsubscript{$\pm$0.012}\) \\
    Hybrid   & \textbf{\(0.294\textsubscript{$\pm$0.019}\)} & \textbf{\(0.213\textsubscript{$\pm$0.032}\)} \\
    \bottomrule
  \end{tabular}
\end{wraptable}

\paragraph{ColBench.}
On \textit{Qwen3-4B}, \method~improves over InfoPO from $0.553$ to $\mathbf{0.685}$ on Pass ($+23.9\%$) and from $0.439$ to $\mathbf{0.600}$ on Succ.\ ($+36.7\%$), increasing the ColBench average from $0.496$ to $\mathbf{0.643}$ ($+29.6\%$). The same pattern holds for \textit{Qwen2.5-7B-Instruct}, where Pass improves from $0.534$ to $\mathbf{0.659}$ ($+23.4\%$) and Succ.\ from $0.426$ to $\mathbf{0.533}$ ($+25.1\%$). For both backbones, the relative gains are larger on the stricter Succ.\ metric, indicating that the improvement extends beyond the more permissive Pass criterion.

\paragraph{$\tau^2$-bench.}
Table~\ref{tab:tau2bench_results} compares \method~with UserRL, InfoPO, and PPO on \textit{Qwen3-4B}. \method~achieves the highest aggregate score, improving from $18.5$ with PPO to $\mathbf{21.5}$ ($+16.2\%$). It achieves the highest scores on Airline and Telecom, whereas InfoPO performs best on Retail.

\subsection{Ablations \& Analysis}
We examine symbolic-user mixing, the EAU-specific intervention design, alternative training algorithms and meta-agent implementations, and intra-run adaptation. Our analysis focuses on capabilities that support autonomous training intervention, rather than treating every editable component as an independently beneficial module.

\paragraph{Ablation on symbolic user.}
Table~\ref{tab:ablation_user} compares LLM, symbolic, and hybrid user simulators on ColBench. For Hybrid, we set the mixing coefficient to $\epsilon=0.5$. The hybrid simulator achieves slightly higher mean Pass and Succ.\ scores than the LLM-only simulator, whereas the symbolic-only variant performs worse. One possible explanation is that mixing more constrained symbolic responses with varied LLM responses preserves interaction diversity while providing more controlled feedback.

\paragraph{Ablation on \method~design.}
Table~\ref{tab:ppo_meta_agent_vs_ours} compares \method~with a generic meta-agent constructed by removing the symbolic user and the skill for intra-run operations over the user, environment, and agent. This baseline receives only a generic instruction to improve RL training. Compared with PPO, it produces mixed results, including a decrease in UserGym performance on \textit{Qwen2.5-7B-Instruct}. \method~outperforms this baseline on both benchmarks and both policy backbones. Because the symbolic user and intervention skill are removed together, this comparison evaluates the combined design rather than their individual effects.

\paragraph{Analysis: Alternative training algorithms and meta-agents.}
Table~\ref{tab:ablation_training_alg_backbone} shows that \method~improves aggregate performance when combined with either UserRL or InfoPO on $\tau^2$-bench. The observed gains are therefore not restricted to the default PPO setup. Replacing the default meta-agent with Codex (\textit{GPT-5.6-Sol}) also improves over InfoPO alone, although the gain is smaller than with the default meta-agent. These results demonstrate compatibility with multiple training and meta-agent configurations, with gains of different magnitudes across the tested configurations.

\begin{table*}[t]
\centering

% ==================== Left table ====================
\begin{minipage}[t]{0.56\textwidth}
\vspace{0pt}
\centering
\captionof{table}{Comparison of PPO without meta-agent, naïve meta-agent without \method, and \method.}
\label{tab:ppo_meta_agent_vs_ours}
\scriptsize
\setlength{\tabcolsep}{3pt}

\begin{tabular}{lc|cc}
\toprule
\multirow{2}{*}{Method} &
UserGym &
\multicolumn{2}{c}{ColBench} \\
\cmidrule(lr){2-2}
\cmidrule(lr){3-4}
& Avg
& Pass & Succ. \\
\midrule

\multicolumn{4}{l}{\textit{Qwen3-4B}} \\

PPO
& 0.525
& 0.284\textsubscript{$\pm$0.046}
& 0.200\textsubscript{$\pm$0.044} \\

Meta-Agent
& 0.574
& 0.317\textsubscript{$\pm$0.038}
& 0.207\textsubscript{$\pm$0.035} \\

\textbf{Ours}
& \textbf{0.622}
& \textbf{0.685\textsubscript{$\pm$0.009}}
& \textbf{0.600\textsubscript{$\pm$0.015}} \\

\midrule

\multicolumn{4}{l}{\textit{Qwen2.5-7B-Instruct}} \\

PPO
& 0.568
& 0.356\textsubscript{$\pm$0.012}
& 0.240\textsubscript{$\pm$0.000} \\

Meta-Agent
& 0.512
& 0.415\textsubscript{$\pm$0.007}
& 0.277\textsubscript{$\pm$0.006} \\

\textbf{Ours}
& \textbf{0.620}
& \textbf{0.659\textsubscript{$\pm$0.017}}
& \textbf{0.533\textsubscript{$\pm$0.018}} \\

\bottomrule
\end{tabular}
\end{minipage}
\hfill
% ==================== Right table ====================
\begin{minipage}[t]{0.41\textwidth}
\vspace{0pt}
\centering
\captionof{table}{
Ablations on training algorithm and meta-agent backbone.
Results are on $\tau^2$-bench.
}
\label{tab:ablation_training_alg_backbone}
\scriptsize
\setlength{\tabcolsep}{4pt}

\begin{tabular}{@{}lcccc@{}}
\toprule
Method & Retail & Airline & Telecom & Avg. \\
\midrule

UserRL
& 16.9 & 7.5 & 13.8 & 13.8 \\

\quad + \method
& \textbf{22.5} & \textbf{15.0} & \textbf{21.9} & \textbf{20.8} \\

\midrule

InfoPO
& \textbf{24.4} & 10.0 & 15.6 & 18.0 \\

\quad + \method
& 20.0 & \textbf{16.3} & \textbf{24.4} & \textbf{21.0} \\

\quad + \method~(Codex)
& 20.0 & 13.8 & 20.0 & 18.8 \\

\bottomrule
\end{tabular}
\end{minipage}

\end{table*}

\begin{figure*}[tbp]
    \centering
    \begin{minipage}[c]{0.48\textwidth}
        \centering
        \includegraphics[width=\linewidth]{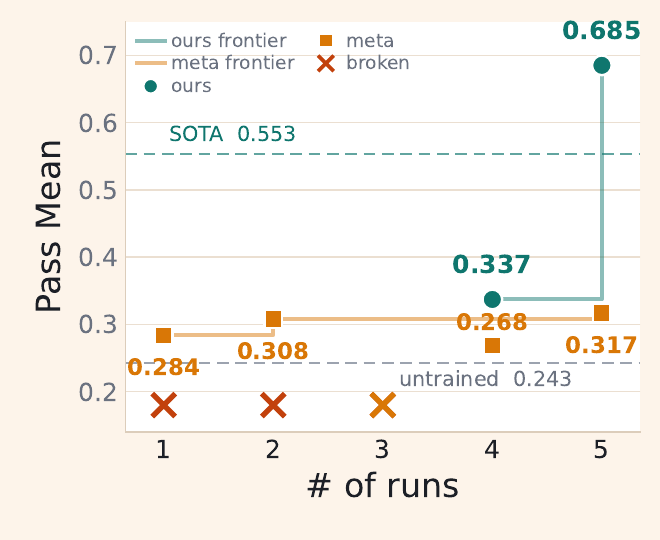}
        \captionof{figure}{
            Best-score frontiers on ColBench for \method~and the generic
            meta-agent. Each frontier tracks the best score observed so
            far and remains flat after unsuccessful or lower-scoring runs.
        }
        \label{fig:frontier}
    \end{minipage}
    \hfill
    \begin{minipage}[c]{0.48\textwidth}
        \centering
        \includegraphics[width=\linewidth]{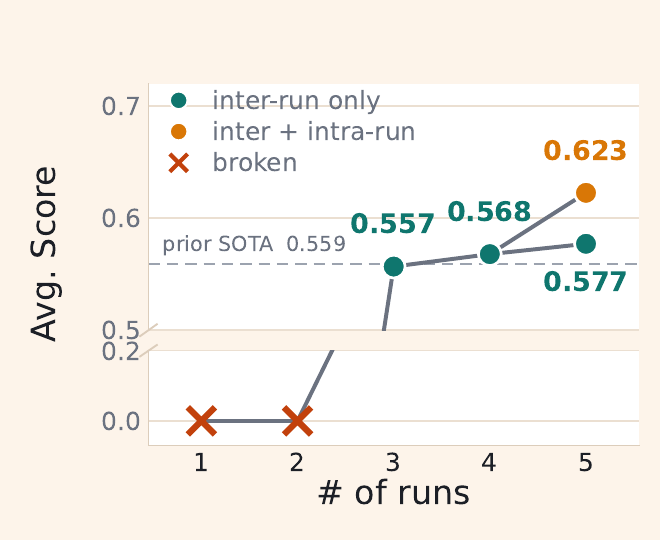}
        \captionof{figure}{
            Run-level progression on UserGym with \textit{Qwen3-4B}.
            The two conditions share experience from the first four runs;
            in run~5, one uses inter-run adaptation only, while the other
            additionally permits intra-run intervention.
        }
        \label{fig:intra_inter_comparison}
    \end{minipage}
\end{figure*}

\paragraph{Analysis: Intra-run exploration and experience reuse.}
We examine how intra-run intervention supports exploration of training modifications and reuse of the resulting experience. On ColBench, \method~proactively restarts its first three runs after detecting severe issues (Figures~\ref{fig:teaser} and~\ref{fig:frontier}). It then explores multiple modifications in run~\textcolor{blue}{4} and carries these observations into run~\textcolor{red}{5}, the second completed run, which surpasses the strongest prior baseline. The generic inter-run meta-agent makes smaller gains over its run sequence. This progression illustrates how intra-run feedback can inform both immediate changes and subsequent training runs.

Figure~\ref{fig:intra_inter_comparison} further examines experience reuse on UserGym. Given the same experience from the first four runs, enabling intra-run intervention in run~\textcolor{orange}{5} yields a larger improvement than inter-run adaptation alone. This supports the benefit of intra-run intervention in the illustrated setting. The curves show progression by run count, rather than an efficiency comparison under matched total compute.

\section{Conclusion}
We introduced \method, a meta-agent framework that jointly evolves the environment,
agent, and user during training. We show that a frontier coding agent can effectively
manage and adapt these components based solely on training feedback, including
trajectories and optimization logs. To provide finer control over user behavior, we
introduce a symbolic user simulator that the meta-agent can modify programmatically.
We further enable intra-run interventions, allowing \method~ to explore and exploit
training modifications more flexibly. Across three user-centric benchmarks, \method~
improves upon prior state-of-the-art performance by up to 36.7\%. This highlights the value of adapting the training system itself.
Future work may investigate post-training open-source models to perform such
co-evolution autonomously.

% \newpage
\section*{Ethics Statement}
Our study concerns RL agents interacting with simulated users in
benchmark environments; it does not recruit human participants or
collect new human-participant data. \method~can modify training code,
rewards, and user-simulator behavior, creating risks of implementation
errors, misleading learning signals, and unintended task-information
disclosure. The documented interface checks constrain admissible updates
but do not guarantee behavioral correctness. Training-time adaptation
is separated from a fixed evaluation protocol, and final-test feedback
is not used to guide interventions or select reported models. These
controls address experimental integrity rather than establish deployment
safety. Performance in simulated benchmarks should not be interpreted
as evidence of safe or reliable behavior with real users.

\section*{Reproducibility Statement}
Section~4 describes the framework and meta-agent workflow.
Section~\ref{sec:experiment} specifies the benchmarks, model
configurations, baselines, evaluation and model-selection protocol,
and training and search limits. Appendix~\ref{sec:knobs} documents
the intervention interfaces, benchmark-specific support, and recorded
intervention outcomes. Additional benchmark and optimization details
are provided in Appendices~\ref{sec:appendix_experiment_details}
and~\ref{sec:additional_implementation_details}. We report standard
deviations across repeated evaluations where available; these do not
by themselves quantify variability across independent repetitions of
the full training and adaptation process. Reproducing the reported
configurations requires access to the external coding-agent and
user-simulator models used in the experiments.

\section*{AI Use Statement}
Generative AI is a functional component of the proposed method and
experimental pipeline. We use coding agents, including
\texttt{Claude Code} and \texttt{Codex}, to inspect training logs and
trajectories and propose or implement changes to the training system.
LLMs also serve as policy models, user simulators, and evaluation
components, as specified in Section~\ref{sec:experiment}.
Beyond these experimental roles, generative AI tools assisted with
manuscript revision, LaTeX preparation, and discussion of experimental
interpretation. The authors retain responsibility for the research
design, implementation, validation of reported results, and all final
interpretations and claims.

\bibliography{ref}
\bibliographystyle{iclr2027_conference}

\newpage
\appendix
% \section{Appendix}
% \section{Knobs Table}
% Qualitative examples of what and how modifications are made during training. Table~\ref{fig:eau_activity} shows the \textit{knobs} and the online modifications for ColBench. Table~\ref{tab:knobs-usergym} demonstrates an example on UserGym.
% \input{tables/knobs_table}
\section{Intervention Interfaces and Run-Level Observations}
\label{sec:knobs}

We distinguish the controls exposed to \method~from the outcomes of
particular interventions. Table~\ref{tab:colbench_knobs} documents the
ColBench interfaces and their recorded use;
Table~\ref{tab:knob_observations} summarizes intervention attempts and
implementation diagnostics; and Table~\ref{tab:usergym_knobs} reports
interface coverage on UserGym. Together, these records describe which
controls were available, what changes were observed, and how the
reported interventions were handled.

\paragraph{Write checks.}
For the documented knob interfaces, write checks enforce declared ranges. Difficulty-increasing updates additionally
require the relevant recent training success rate to reach $0.80$.
In the ColBench records, four proposed updates to
\texttt{max\_steps} were rejected under this rule before the run
qualified. These checks constrain admissible writes; they do not
guarantee that an update improves performance or achieves its intended
behavioral effect.

\subsection{ColBench: Interface Usage}

Table~\ref{tab:colbench_knobs} covers two analyzed runs of $624$ training
steps each. We distinguish controls that were adjusted, controls that
remained fixed or were not exercised, and interfaces with restricted
implementation scope. An unused control is not a failed intervention,
while an exposed control need not govern every behavior suggested by
its name.

\begin{table}[htbp]
\centering
\small
\setlength{\tabcolsep}{4pt}
\renewcommand{\arraystretch}{1.12}

\caption{
    ColBench interfaces and their use in the two analyzed runs.
    ``Not exercised'' refers to these runs only; ``unavailable'' denotes
    unsupported functionality in the documented implementation.
    A dash in the range column means that no range is specified or applicable.
}
\label{tab:colbench_knobs}

\begin{tabularx}{\linewidth}{
    @{}
    >{\raggedright\arraybackslash}p{0.43\linewidth}
    >{\centering\arraybackslash}p{0.13\linewidth}
    >{\raggedright\arraybackslash}X
    @{}
}
\toprule
\textbf{Knob}
& \textbf{Range}
& \textbf{Recorded use / scope} \\
\midrule

\multicolumn{3}{@{}l}{\textit{Environment}} \\

\texttt{max\_steps}
& $3$--$60$
& Adjusted and reverted. \\

\texttt{min\_questions}
& $0$--$3$
& Counter repaired; parameter adjusted. \\

\texttt{human\_response\_char\_limit}
& ---
& Not exercised. \\

\addlinespace[3pt]
\multicolumn{3}{@{}l}{\textit{User (symbolic)}} \\

\texttt{facts\_per\_reply}
& $1$--$3$
& Adjusted and reverted. \\

\texttt{match\_threshold}
& $[0,1]$
& Multiple settings examined. \\

\texttt{volunteer\_interval}
& $0$--$10$
& Adjusted and reverted. \\

\texttt{max\_reveals}
& $0$--$20$
& Does not bound repeated disclosure of known facts. \\

\texttt{reveal\_return\_shape}
& $\{\mathrm{T},\mathrm{F}\}$
& Filters \texttt{shape}, not \texttt{return}, facts. \\

\addlinespace[3pt]
\multicolumn{3}{@{}l}{\textit{Reward}} \\

\texttt{per\_gym\_scale}
& $[0,2]$
& Not exercised. \\

\texttt{turn\_reward\_clip}
& $[-1,2]$
& Not exercised. \\

\texttt{predicate\_weight}
& ---
& Unavailable; no ColBench predicate library. \\

\addlinespace[3pt]
\multicolumn{3}{@{}l}{\textit{User (mixture / persona)}} \\

\texttt{user\_strategy\_mix}
& $[0,1]$
& Fixed at $0.5$. \\

\texttt{user\_persona}
& ---
& Unavailable on ColBench; travel only. \\

\addlinespace[3pt]
\multicolumn{3}{@{}l}{\textit{Agent}} \\

\texttt{entropy\_coeff}
& $[0,0.05]$
& Adjusted and reverted. \\

\bottomrule
\end{tabularx}
\end{table}

\FloatBarrier

\subsection{ColBench: Intervention Attempts and Diagnostics}

The run summary records $15$ knob writes, two measured harms, and three
subsequently retracted attributions. It also records trajectory-based
diagnoses of deadlock, a reward-hacking channel, and template collapse.
Table~\ref{tab:knob_observations} groups related observations rather than
listing every write separately. It distinguishes the observed behavior
from the reported response: a code repair, a rollback, an inconclusive
perturbation, or a diagnostic finding without a documented follow-up
repair.

% Author check before submission:
% Identify the data source and timing of the Pass / flow-success
% measurements below. Do not relabel them as training-only feedback
% unless the execution records establish this.
%
% The original summary does not define sigma or its estimation procedure.
% Raw score changes are retained without the original >2-sigma claim.

\begin{table}[htbp]
\centering
\footnotesize
\setlength{\tabcolsep}{4pt}
\renewcommand{\arraystretch}{1.13}

\caption{
    Intervention attempts and implementation diagnostics in the two
    ColBench runs. Reported score changes are local observations, not
    controlled estimates of each intervention's contribution to final
    performance. A missing follow-up entry does not establish that a
    problem persisted throughout all subsequent runs.
}
\label{tab:knob_observations}

\begin{tabularx}{\linewidth}{
    @{}
    >{\raggedright\arraybackslash}p{0.23\linewidth}
    >{\hsize=1.18\hsize
      \linewidth=\hsize
      \raggedright\arraybackslash}X
    >{\hsize=0.82\hsize
      \linewidth=\hsize
      \raggedright\arraybackslash}X
    @{}
}
\toprule
\textbf{Control / issue}
& \textbf{Observed behavior}
& \textbf{Reported response / status} \\
\midrule

\texttt{max\_steps}
& Reducing $10\rightarrow6$ lowered Pass by $0.246$;
  reverting increased it by $0.177$.
& Reverted. The perturbation-and-reversal pattern concerns sensitivity
  to the interaction budget. \\

\addlinespace[4pt]

\texttt{min\_questions}
& The counter omitted turns that both asked and answered a question,
  deadlocking $41\%$ of episodes.
& Code repair reported. Subsequent $1\rightarrow2$ adjustment had no
  measured effect; $2\rightarrow3$ was harmful. \\

\addlinespace[4pt]

\texttt{facts\_per\_reply}
& Increasing $1\rightarrow3$ produced reference-solution disclosure
  in $837$-character replies.
& Reverted. The initial reward attribution was not retained after
  accounting for the reported gap band. \\

\addlinespace[4pt]

\texttt{match\_threshold}
& The sequence $0.9/0.6/0.35/0.9$ produced no change exceeding the
  reported noise floor of $0.228$.
& Inconclusive at the reported noise level. \\

\addlinespace[4pt]

\texttt{volunteer\_interval}
& Reducing $10\rightarrow5$ increased flow success while the question
  rate fell from $40\%$ to $26\%$.
& Reverted; the rise in flow success was interpreted as masking
  reduced questioning. \\

\addlinespace[4pt]

\texttt{max\_reveals}
& Re-disclosing an already revealed fact did not consume the budget,
  leaving repeated disclosure unrestricted by this control.
& Interface limitation documented; no corresponding repair is
  specified in the run summary. \\

\addlinespace[4pt]

\texttt{reveal\_return\_shape}
& The filter targeted \texttt{kind=="shape"} ($8\%$ of facts), whereas
  the observed leak involved \texttt{kind=="return"} ($18\%$).
& Type mismatch documented; no corrective edit is specified in
  the run summary. \\

\addlinespace[4pt]

\texttt{entropy\_coeff}
& Increasing $0\rightarrow0.004$ raised \texttt{actor/entropy} by
  $50\%$ while behavioral diversity decreased.
& Reverted. The observed effect lag exceeded the $10$-step
  verification window. \\

\bottomrule
\end{tabularx}
\end{table}

\FloatBarrier

\paragraph{Interpreting the intervention record.}
The \texttt{min\_questions} example separates a code repair from later
parameter search: repairing a counter does not imply that further
increasing its threshold is beneficial. Similarly, the disclosure and
volunteering examples show why a reward or success metric must be
interpreted alongside the interaction behavior. Conversely, the
\texttt{max\_reveals} and \texttt{reveal\_return\_shape} findings identify
limitations of the exposed controls, not successful adaptations.
Trajectory inspection and knob adjustment thus address different parts
of the adaptation process: diagnosing a problem and attempting a change.

\paragraph{Fixed mixture.}
The symbolic--LLM mixing coefficient remained at $0.5$, while the two
simulator arms differed by $0.29$ in success rate. The blended metric
therefore combines distinct interaction conditions. These runs do not
provide evidence for gains from dynamically adapting the mixture,
and differences between the simulator arms should not be conflated
with effects of changing the mixing coefficient.

\subsection{UserGym: Benchmark-Specific Interface Coverage}

Table~\ref{tab:usergym_knobs} distinguishes supported controls from
unavailable ones in the documented UserGym implementation. A checkmark
indicates interface support, not that the control was used or improved
performance. Coverage is benchmark-specific, and the mixture-selector
mapping imposes additional restrictions beyond the declared knob range.

\begin{table}[htbp]
\centering
\small
\setlength{\tabcolsep}{3pt}
\renewcommand{\arraystretch}{1.12}

\caption{
    UserGym interface coverage in the documented implementation.
    $\checkmark$ indicates support and --- indicates unavailable
    functionality. These entries describe implementation coverage
    rather than intervention outcomes.
}
\label{tab:usergym_knobs}

\begin{tabularx}{\linewidth}{
    @{}
    >{\raggedright\arraybackslash}X
    ccccc
    @{}
}
\toprule
\textbf{Knob (range)}
& \textbf{travel}
& \textbf{tau}
& \textbf{persuade}
& \textbf{turtle}
& \textbf{function} \\
\midrule

\multicolumn{6}{@{}l}{\textit{Environment}} \\

\texttt{max\_steps} ($3$--$60$)
& $\checkmark$ & $\checkmark$ & $\checkmark$
& $\checkmark$ & $\checkmark$ \\

\texttt{search\_failure\_interval} ($1$--$50$)
& $\checkmark$ & --- & --- & --- & --- \\

\texttt{elicitation\_interval} ($1$--$20$)
& $\checkmark$ & --- & --- & --- & --- \\

\texttt{noise\_choice\_number} ($0$--$30$)
& $\checkmark$ & --- & --- & --- & --- \\

\texttt{wrong\_choice\_number} ($0$--$30$)
& $\checkmark$ & --- & --- & --- & --- \\

\texttt{success\_threshold}
& --- & --- & --- & $\checkmark$ & --- \\

\texttt{max\_repeated\_probes}
& --- & --- & --- & --- & $\checkmark$ \\

\addlinespace[3pt]
\multicolumn{6}{@{}l}{\textit{Reward}} \\

\texttt{default\_scale} $[0,2]$
& $\checkmark$ & $\checkmark$ & $\checkmark$
& $\checkmark$ & $\checkmark$ \\

\texttt{per\_gym\_scale} $[0,2]$
& $\checkmark$ & $\checkmark$ & $\checkmark$
& $\checkmark$ & $\checkmark$ \\

\texttt{turn\_reward\_clip} $[-1,2]$
& $\checkmark$ & $\checkmark$ & $\checkmark$
& $\checkmark$ & $\checkmark$ \\

\texttt{predicate\_weight} $[-0.5,0.5]$
& --- & --- & --- & --- & $\checkmark^{a}$ \\

\addlinespace[3pt]
\multicolumn{6}{@{}l}{\textit{User}} \\

\texttt{user\_strategy\_mix} $[0,1]$
& $\checkmark$ & ---$^{b}$ & ---$^{b}$ & ---$^{b}$ & ---$^{b}$ \\

\texttt{user\_persona} ($\leq8$)
& $\checkmark$ & --- & --- & --- & --- \\

\addlinespace[3pt]
\multicolumn{6}{@{}l}{\textit{Agent}} \\

\texttt{entropy\_coeff} $[0,0.05]$
& \multicolumn{5}{c@{}}{
    Shared across gyms; change per write $\leq0.004$.
} \\

\bottomrule
\end{tabularx}

\vspace{3pt}
\begin{minipage}{\linewidth}
\footnotesize

\textsuperscript{a}~\texttt{function} is the only listed gym with a
predicate library (five \texttt{probe\_*} predicates);
\texttt{predicate\_weight} is rejected elsewhere.

\par\smallskip

\textsuperscript{b}~The mixture selector is keyed by
\texttt{travel}/\texttt{colbench}/\texttt{tau2}. The other four UserGym
keys resolve to \texttt{None}, so they cannot be split into symbolic
and LLM arms through this interface; in particular, \texttt{tau} does
not match \texttt{tau2}. This describes the recorded implementation,
not a benchmark requirement.

\end{minipage}
\end{table}

\FloatBarrier
\begin{figure}[t]
    \centering
    \includegraphics[width=0.99\linewidth]{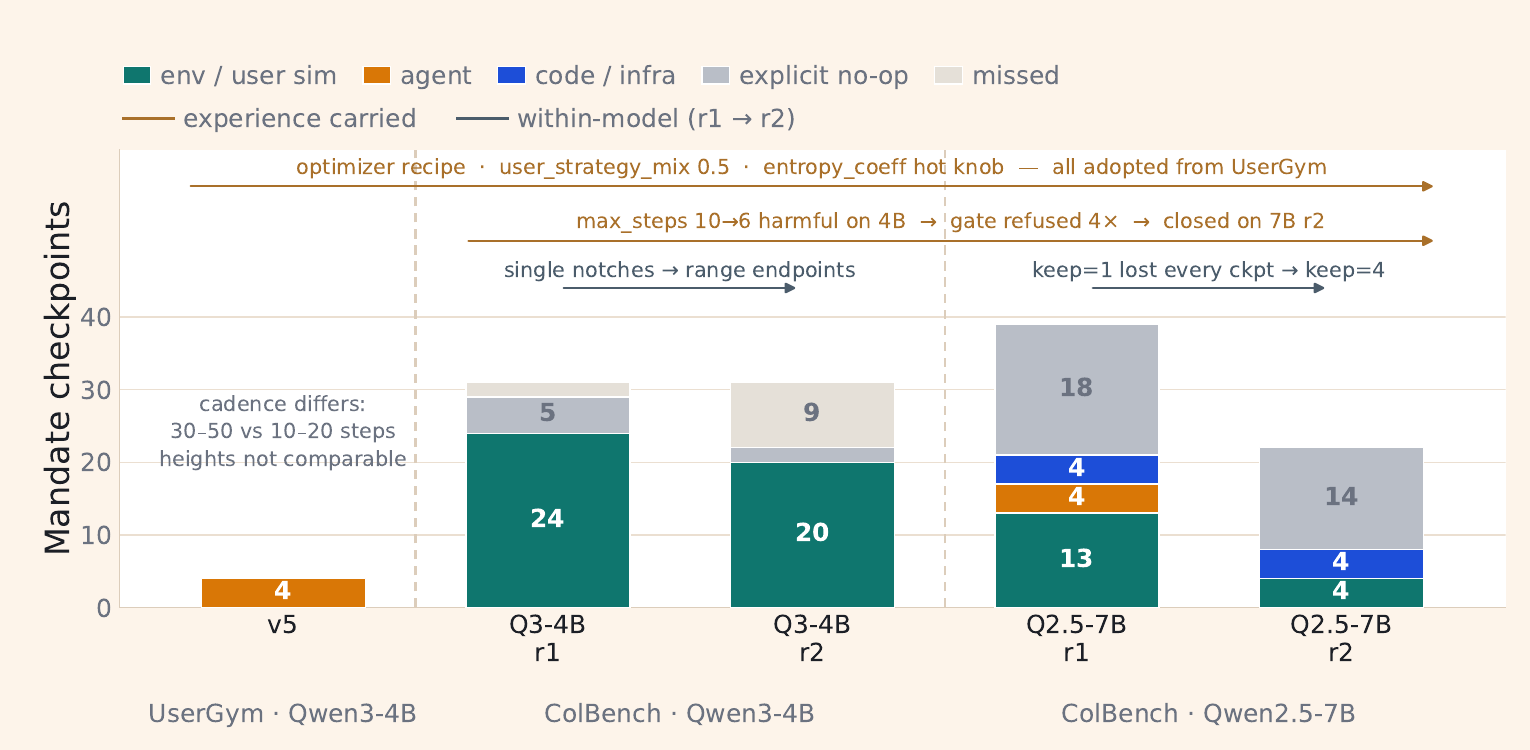}
    \caption{Experience flow from UserGym, \textit{Qwen3-4B} to ColBench, \textit{Qwen2.5-7B}.}
    \label{fig:eau_activity}
    \vspace{-10px}
\end{figure}
\section{Details on Experiments}
\label{sec:appendix_experiment_details}
\paragraph{Baselines.}
We compare EAU against both prompting-based and RL-based baselines.
For prompting, we include ReAct~\citep{yao2022react}, which interleaves reasoning with actions, and Reflexion~\citep{shinn2023reflexion}, which improves agent behavior through verbal feedback and self-reflection.
For RL training, we consider RAGEN~\citep{wang2025ragen}, a general multi-turn agent RL framework; Search-R1~\citep{jin2025search}, which trains agents through reinforcement learning over multi-step search trajectories; UserRL~\citep{qian2025userrl}, which targets interactive user-centric agents; and InfoPO~\citep{kong2026infopo}, which introduces information-driven optimization for user-agent interaction.
We additionally include vanilla PPO as a standard RL baseline to isolate the gains from our co-evolution framework.
For UserGym and ColBench, all baseline results except PPO are taken from \citet{kong2026infopo}; PPO and EAU are evaluated under our implementation.
\paragraph{Benchmarks.} We evaluate our method on three representative user-centric benchmarks, i.e., \textbf{ColBench}~\citep{zhou2025sweet}, \textbf{UserGym}~\citep{qian2025userrl}, and \textbf{$\tau^2$-bench}~\citep{yao2024tau,barres2025tau2}. ColBench evaluates collaborative artifact creation, where the agent works with an LLM-simulated collaborator that has privileged access to a hidden reference artifact. We evaluate \method~on the \emph{Backend Programming} subset, where the agent implements a Python function from an incomplete specification, and is evaluated with hidden unit tests. UserGym is a suite of eight Gymnasium environments unified under a single tool interface, spanning preference elicitation, intent clarification, entity guessing, lateral-thinking puzzles, rule induction, policy-constrained tool use, persuasion, and search. Each episode pairs a rule-based task with an LLM-simulated user who holds the ground-truth intent and reveals it only when queried. Because the initial request is systematically underspecified, the agent must repeatedly trade off asking the user against committing to an action. We report the environment reward averaged within and across gyms. Specifically, we exclude the SearchGym for evaluation due to the cost of APIs. It requires the agent to conduct online search using a charged api. Notably, this does not affect the RL training, since it's an OOD gym solely for evaluation. $\tau^2$-bench casts customer service as a dual-control problem in which both the agent and the simulated user can act on a shared world state. Unlike prior benchmarks where the user is a passive information source, the user here has its own tools, so the agent must verbally guide the user through actions it cannot perform itself while complying with a domain policy. We evaluate on the retail, airline, and telecom domains, and score a task as solved only when the resulting environment state matches the annotated target and all required information has been conveyed to the user.

\section{Additional Implementation Details}
\label{sec:additional_implementation_details}
We show the hyperparameters in Table~\ref{tab:hyperparameters}.
\begin{table}[ht]
\centering
\caption{Key hyperparameters for training on UserGym, ColBench, and $\tau^2$-Bench environments.}
\label{tab:hyperparameters}
\begin{tabular}{@{}lccc@{}}
\toprule
\textbf{Parameter} 
& \textbf{UserGym} 
& \textbf{ColBench} 
& \textbf{$\tau^2$-Bench} \\
\midrule

% \multicolumn{4}{l}{\textit{Info Gain Configuration}} \\
% Intrinsic weight ($\beta_0$)         & 0.5              & 0.3              & 0.1 \\
% KL batch size                        & 4                & 4                & 2 \\
% Gate temperature                     & 0.5              & 0.5              & 0.05 \\
% \midrule

\multicolumn{4}{l}{\textit{Data Configuration}} \\
Train batch size                     & 64               & 64               & 32 \\
Max prompt length                    & 1152             & 2048             & 8192 \\
Max response length                  & 8192             & 8192             & 16384 \\
\midrule

\multicolumn{4}{l}{\textit{Actor Configuration}} \\
Learning rate                        & $3\times10^{-7}$  & $3\times10^{-7}$  & $1\times10^{-6}$ \\
PPO minibatch size                   & 16               & 16               & 16 \\
PPO microbatch size per GPU          & 4                & 4                & 2 \\
KL loss coef                         & 0.001            & 0.001            & 0.001 \\
Entropy coefficient                  & 0.001            & 0.001            & 0.001 \\
\midrule

\multicolumn{4}{l}{\textit{Rollout Configuration}} \\
Rollout engine                       & SGLang           & SGLang           & SGLang \\
Log prob micro batch size per GPU    & 4                & 4                & 2 \\
Parallel rollouts ($n$)              & 5                & 5                & 5 \\
Max turns                            & 16               & 10               & 50 \\
Response length (per turn)           & 1024             & 1024             & 1024 \\
GPU memory utilization               & 0.5              & 0.5              & 0.5 \\
\midrule

\multicolumn{4}{l}{\textit{Training Configuration}} \\
GPUs per node                        & 4                & 4                & 4 \\
Total epochs                         & 2                & 2                & 2 \\
\bottomrule
\end{tabular}
\end{table}
\section{Additional Analysis}
\paragraph{Observations from the main results.}
We observe that with a strong model, i.e., \textit{Qwen3-4B}, \method~is able to achieve generally better performance. On benchmarks involving capabilities with which the model is more familiar---namely coding (ColBench) rather than tool use (UserGym)---\method~ can achieve relatively greater performance gains. This may suggest that \method~ is more adept at eliciting better interaction behavior from the model than at overcoming its intrinsic capability limitations.

\end{document}